\documentclass[journal]{IEEEtran}
\IEEEoverridecommandlockouts
\usepackage{graphicx}
\usepackage{etoolbox}
\usepackage{nicematrix}
\usepackage{tikz}
\usepackage{xspace}
\usepackage{bm}
\usepackage{unicode}
\usepackage[flushleft]{threeparttable}
\usepackage{makecell}
\usepackage{enumitem}
\usepackage{mwe}
\usepackage[fontsize=10.5pt]{fontsize}
\usepackage{hyperref}
\usepackage{academicons}
\usepackage[column=0]{cellspace}

\newcommand{\etal}{\textit{et al.}}
\begin{document}

\title{Balancing Privacy and Progress in Artificial Intelligence: Anonymization in Histopathology for Biomedical Research and Education}

\author{Neel Kanwal$^1$*\href{https://orcid.org/0000-0002-8115-0558}{\includegraphics[scale=0.01]{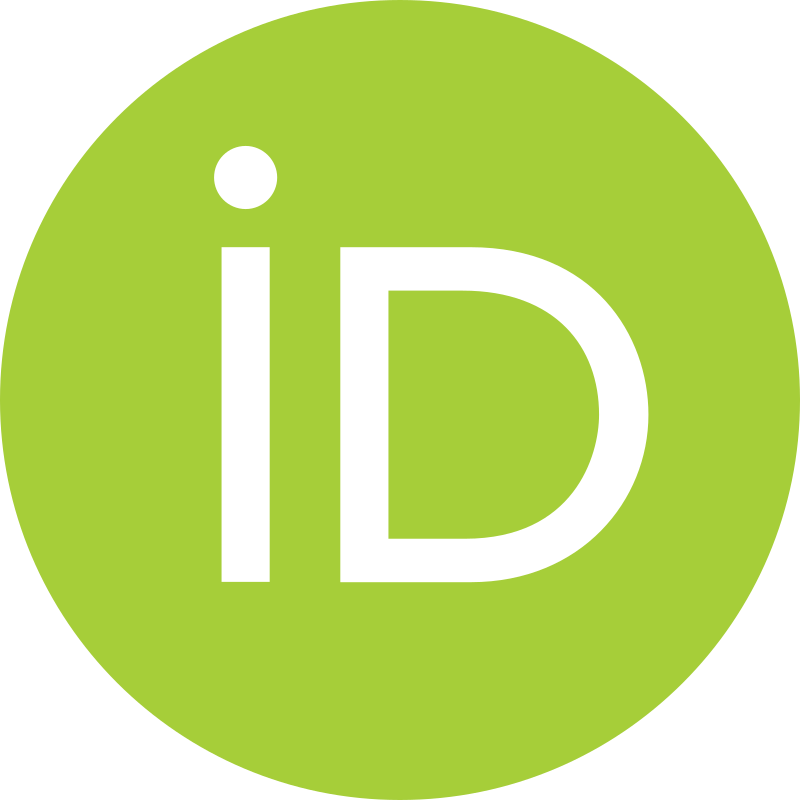}}, Emiel A.M. Janssen$^{2,3}$, Kjersti Engan$^1$\href{https://orcid.org/0000-0002-8970-0067}{\includegraphics[scale=0.01]{orcid.png}}\\
$^1$Department of Electrical Engineering and Computer Science, University of Stavanger, Norway \\
$^2$Department of Chemistry, Bioscience and Environmental Engineering, University of Stavanger, Norway\\
$^3$Department of Pathology, Stavanger University Hospital, Stavanger, Norway\\
*Corresponding author: neel.kanwal@uis.no}

\maketitle
\begin{abstract}
The advancement of biomedical research heavily relies on access to large amounts of medical data. In the case of histopathology, Whole Slide Images (WSI) and clinicopathological information are valuable for developing Artificial Intelligence (AI) algorithms for Digital Pathology (DP). Transferring medical data "as open as possible" enhances the usability of the data for secondary purposes but poses a risk to patient privacy. At the same time, existing regulations push towards keeping medical data "as closed as necessary" to avoid re-identification risks. Generally, these legal regulations require the removal of sensitive data but do not consider the possibility of data linkage attacks due to modern image-matching algorithms. In addition, the lack of standardization in DP makes it harder to establish a single solution for all formats of WSIs. These challenges raise problems for bio-informatics researchers in balancing privacy and progress while developing AI algorithms. This paper explores the legal regulations and terminologies for medical data-sharing. We review existing approaches and highlight challenges from the histopathological perspective. We also present a data-sharing guideline for histological data to foster multidisciplinary research and education.  

\end{abstract}


\section{Introduction}
% Background on the importance of medical data sharing for research and education
%% Importance of sharing in digital pathology
% Incidents where Improper sharing lead to losses.
% Overview of the challenges related to privacy and regulations in histopathology, FAIR principles not fully applicable
% lack of standardization, hard to balance privacy and progress
% Thesis statement: Anonymization of whole slide images is a promising solution to preserve privacy and enable medical data sharing for research and education.
% \vspace{-0.5em}
\vspace{0.5em}
In recent years, the adoption of data management systems and cloud technologies has made the healthcare industry data-rich. Sharing medical data is essential for fostering collaboration and accelerating scientific progress in biomedical research and education. However, biomedical research has long been hindered by limited access to medical data. For instance, Digital Pathology (DP) has considerable potential, where Artificial Intelligence (AI) algorithms may help pathologists save considerable amounts of time, make more precise diagnoses, and provide secondary opinions~\cite{kanwal2023vision}. Also, for education, pathologists can benefit from understanding broad patterns of clinical conditions and rare diseases by pooling large amounts of data from different institutions worldwide. These large datasets are also advantageous in developing more robust AI algorithms~\cite{kanwal2023vision,tabatabaei2023wwfedcbmir}. Developing AI algorithms for diagnosis and treatment also requires a computational infrastructure, which is usually unavailable at healthcare institutions, making it necessary to transport medical data outside the premises~\cite{tabatabaei2023wwfedcbmir,wang}. Besides the challenges in preparing large datasets, sharing data in interdisciplinary research raises privacy concerns. Although several regional and local regulations provide a general framework for de-identifying sensitive information~\cite{el2011systematic}, they compromise the vast usability of medical data for different scenarios, such as epidemiology, prognosis follow-ups, etc.

Improper data sharing may lead to severe repercussions for organizations, such as damage to reputation and hefty fines. Various incidents of inadvertent data exposure and compliance failure have occurred in the past, such as a UK-based telecom company, TalkTalk, which faced a data breach due to a database injection attack exposing the personal information of 157,000 customers and facing a fine of £0.4 million~\cite{talktalk}. Similarly, Anthem, one of the largest healthcare insurance companies, failed to comply with local regulations in protecting the health data of nearly 79 million people and received a penalty of \$16 million~\cite{anthem}. In 2017, the University of Rochester Medical Center (URMC) in New York faced a data breach from unencrypted flash drives containing information of 3,400 active patients~\cite{urmc}. Their failure to place adequate protection measures caused the imposition of a \$3 million fine. Among all incidents of data breaches reported between 2005 and 2019, the healthcare industry faced the highest number of breaches~\cite{seh2020healthcare}. Therefore, it has become essential for medical data custodians to implement robust security measures and fully comply with regulations for protecting sensitive information before sharing it for multidisciplinary research.

De-identifying medical data may mitigate the risk of drastic consequences such as data breaches, insider threats, ransomware attacks, and other security issues~\cite{halim2021effective}. Despite de-identification, technological advancements pose non-trivial challenges, such as the risk of re-identifying patients, possibly in combination with other available databases~\cite{maritsch2022data,sweeney2013matching}. In the case of histopathology, Whole Slide Images (WSIs) without metadata may still lead to identity disclosure attacks through image-matching algorithms~\cite{holub2023privacy}. With relatively little computational effort, modern image-matching algorithms can extract features from the tissue in the original WSI and identify the hospital or lab based on specific staining and possibly by combining information about the hospital with the uniqueness of the disease. These data linkage attacks make it hard to balance privacy and progress by keeping data \emph{as open as possible} for authorized use and \emph{as close as necessary} for unauthorized use. 

Existing anonymization  frameworks~\cite{andrew2023anonymization,mehta2022improved,welten2022privacy} (discussed later in section~\ref{relatedwork}) and FAIR (Findable, Accessible, Interoperable, and Reusable) principles~\cite{queralt2022applying} are multifaceted and do not provide a straightforward solution for DP. It is also due to the fact that the DP domain itself lacks standardization in digitization and practices for maintaining clinical information, and WSIs with different data formats have a different structure for stored metadata. Thus, a single solution is not applicable to histopathological data from different sources. Moreover, confusion arises from the widespread usage of different terminologies like encryption/coding, anonymous, pseudonymization, and de-identified data for privacy-preserving purposes. In this article, we will explain legal regulations and these terminologies and definitions under legal frameworks. 
This paper reviews existing approaches and challenges in their adoption in histopathology.
We also provide guidelines and ethical considerations for exchanging histopathological data for research and future directions for facilitating cloud-based DP services.

\section{Regulations for Medical Data Sharing}
% Definition of anonymization and its importance in medical data sharing.
% Key concepts and principles for anonymizing health data while ensuring it remains suitable for meaningful analysis.

% Current privacy approaches for medical text data and their limitations.
\vspace{0.5em}
Medical data exchange must comply with local, national, and international laws, which provide a minimal framework to facilitate the usage of medical data for secondary purposes. In order to build an infrastructure for sharing medical data, it is important to understand the legal and regulatory aspects.

\subsection{Legal and Regulatory Aspects}
Though data protection rules and jurisdiction differ from country to country, they all share the similar goal of protecting patients' confidentiality and privacy. Europe and the United States (U.S.) have different laws defining "identifiable" and "non-identifiable" medical data. However, the definitions do not consider recent technological advancements. These regulations only provide minimal safeguards to establish a secure environment that will ensure legal certainty.

In the U.S., the Health Insurance Portability and Accountability Act (HIPAA) of 1996 regulates the use of protected health information~\cite{nosowsky2006health}. HIPAA requires businesses and the healthcare industry to avoid disclosing health data without patients' consent. HIPAA only applies to healthcare providers and data hosting companies in the U.S., which means that patients' personal data may not be adequately protected when shared with organizations outside of the U.S. In the European Union (E.U.), the General Data Protection Regulation (GDPR) regulates the handling of sensitive personal data and defines provisions for medical information~\cite{GDPRwhitepaper}. The GDPR sets the legal bar for the minimal requirements for all E.U. countries on how to exchange medical data. GDPR applies to all organizations that process the personal data of E.U. residents, regardless of where the organization is located. HIPAA and GDPR are complex, leading to confusion and unintentional compliance issues.

GDPR is lenient compared to HIPAA and imposes small fines to encourage hospitals to protect patient information instead of bankrupting them. Under GDPR, the data subject owns the data, while under HIPAA, the covered entity owns the data. Both regulations give the right to patients to amend, inspect, or restrict the use of their data~\cite{pesapane2021legal}. These regulations mandate that healthcare institutions have robust security measures and data management policies, making it hard for hospitals to share medical data for multidisciplinary research and ultimately affecting the innovation of AI-based healthcare technologies.

% \vspace{-1em}
\subsection{Understanding Terminologies under Regulations}
% Direct, Indirect, and Quasi Identifiers
% encryption, deidentified, pseudonymized, anonymized

Encryption, de-identification, pseudonymization, and anonymization techniques are all used to protect sensitive data. While some of them are often confused with each other, they hold key differences under legal regulations. These privacy-preserving measures are applied to all identifiers in the datasets, which can be broadly divided into two categories: direct identifiers and quasi-identifiers~\cite{el2015anonymising}. Direct identifiers are attributes, such as names, email addresses, phone numbers, and social insurance numbers, that enable direct identification of individuals. In contrast, quasi-identifiers are characteristics that can be used to indirectly infer someone's identity and include ethnicity, date of birth, date of death, date of a visit to a clinic, and the postal code of the address. Protecting both the direct identifiers and the quasi-identifiers is, therefore, crucial.

\begin{itemize}  
    \item Encryption methods hide information in identifiers using cryptography to avoid unauthorized access~\cite{xu2022privacy}. Data encryption is usually applied when storing data in a database, and protection is lost when the data is decrypted. By design, encryption is considered an appropriate security measure for privacy.
    \item Pseudonymization is the process of replacing direct identifiers with pseudonyms. Encryption is sometimes used to create pseudonyms from patient identifiers. These pseudonyms can only be linked to specific individuals by authorized entities (via a key).
    \item Data de-identification refers to removing all direct and quasi-identifiers from personal data. De-identification aims to make it difficult to re-identify individuals. The term \emph{de-identification} is often interchangeably used for anonymization in the US but carries subtle differences under GDPR~\cite{chevrier2019use,garfinkel2015identification}.
    \item Anonymization, on the other hand, involves transforming the data in such a way that the identity of individuals cannot be determined (only identifiable by disproportionate effort and time). Anonymization is considered irreversible and more secure than simpler de-identification.
\end{itemize}
 
Pseudonymization and anonymization are two popular strategies employed when data leaves the institution's premises. Both HIPAA and GDPR differ in their approach to the terms anonymization and pseudonymization. HIPAA uses the term "de-identification" rather than anonymization or pseudonymization, where de-identification is not the same as anonymization in GDPR, as de-identified data may still contain some indirect information~\cite{chevrier2019use}. Moreover, GDPR does not recognize de-identified data as a separate category. GDPR mentions pseudonymization as a core technique for data protection and anonymization as a privacy-enhancing technique~\cite{pesapane2021legal}. Pseudonymized data can be shared, provided that the correct data fields are pseudonymized. Anonymized data, as opposed to pseudonymized data, is no longer designated personal information by the GDPR and is not further subject to data protection legislation. Though anonymization offers enhanced security over pseudonymization, it reduces the utility of medical data. In short, pseudonymization is a more flexible technique that allows data linkage from different sources. At the same time, anonymization is a more secure technique but, unfortunately, limits the usefulness of the data for research purposes.

\section{Privacy-Preserving Frameworks}
\label{relatedwork}
% Description of an anonymization framework for whole slide images
% Discussion of the key components of the framework, including data de-identification, data masking, and data encryption
% Examples of how the framework can be applied in medical research and education
\vspace{0.5em}
Data is a valuable resource in this age of information explosion, but resources like medical data are often at a higher risk of privacy leakage. Several methods (see review~\cite{el2011systematic}) have attempted to re-identify patient information from public datasets, while others (see review~\cite{majeed2020anonymization}) have added layers of security to prohibit data linkage. Unsurprisingly, the development of privacy-preserving frameworks has been an emerging research topic with substantial literature. These frameworks can be broadly categorized as traditional anonymization, cryptographic, and distributed computation techniques.

% Some recent research works have attempted to exploit and re-identify patient data from public datasets, while many others have attempted to add layers of security to prohibit data linkage.

%% Traditional Anonymization

Traditional anonymization can be further grouped into simpler anonymization and advanced anonymization techniques. Simpler anonymization techniques alter or remove explicit identifiers to reduce the risk of unintended disclosure. Some popular techniques in this group are generalization, suppression, and perturbation~\cite{andrew2023anonymization,majeed2020anonymization}. Generalization refers to replacing specific values with broader categories to prevent individual identification. For example, instead of reporting an exact age, age ranges can be used. Suppression involves removing certain variables or data points entirely from the dataset. Perturbation entails adding random noise or slightly altering values to protect privacy while preserving statistical properties. The second group, advanced anonymization techniques, is more sophisticated and aims to reduce disclosure risks by grouping quasi-identifiers in such a way that they remain indistinguishable.
\emph{K-anonymity} is a pioneering statistical disclosure control technique that aims to ensure that each record in a single dataset is indistinguishable from at least k-1 other records with respect to certain identifying attributes. Other derived methods in this group, such as t-closeness, l-diversity, and others, compensate for the cons of k-anonymity but do not foresee the problem of data linkage to other available databases~\cite{mehta2022improved,rajendran2017study}. Andrew \etal~\cite{andrew2023anonymization} proposed a protocol for multiple data owners to tackle internal and external identity disclosure problems. Their approach was based on k-anonymity groups and greedy heuristics to allocate patients to groups. However, they did not consider membership disclosure or similarity attacks between groups. Recently, Mehta \etal~\cite{mehta2022improved} proposed an improved l-diversity approach for scalable privacy solutions. Their approach used a clustering-based technique to reduce information loss. Unfortunately, the traditional anonymization approaches are not applicable for sanitizing large data enclaves (with multiple data owners) due to utility loss and the possibility of disclosure and inference attacks.

%% Cryptographic
\emph{Cryptographic techniques} use encryption as a fundamental to ensure authorized use only. Attribute-based Encryption (ABE) and Homomorphic Encryption (HE) are popular techniques used for different purposes. ABE is used for one-to-many data distribution where fine-grained access control is established (using a key) between the patient and data users. Xu \etal~\cite{xu2022privacy} developed ABE to grant control of the data to the owner. Their revocable mechanism aimed for flexible data control with cross-hospital expertise. Conversely, HE allows computations on encrypted data without decrypting it. Kocabas \etal~\cite{kocabas2015utilizing} explored implementation aspects of HE for medical cloud computing. Later, Carpov \etal~\cite{carpov2016practical} developed a mobile application to offload medical data over the cloud for analysis. Both of these works have performance disadvantages in terms of complexity and scalability for image data, and there is a need for more efficient cryptographic techniques for histopathology.

\begin{figure*}[ht!]
    \centering
    \includegraphics[width=18cm]{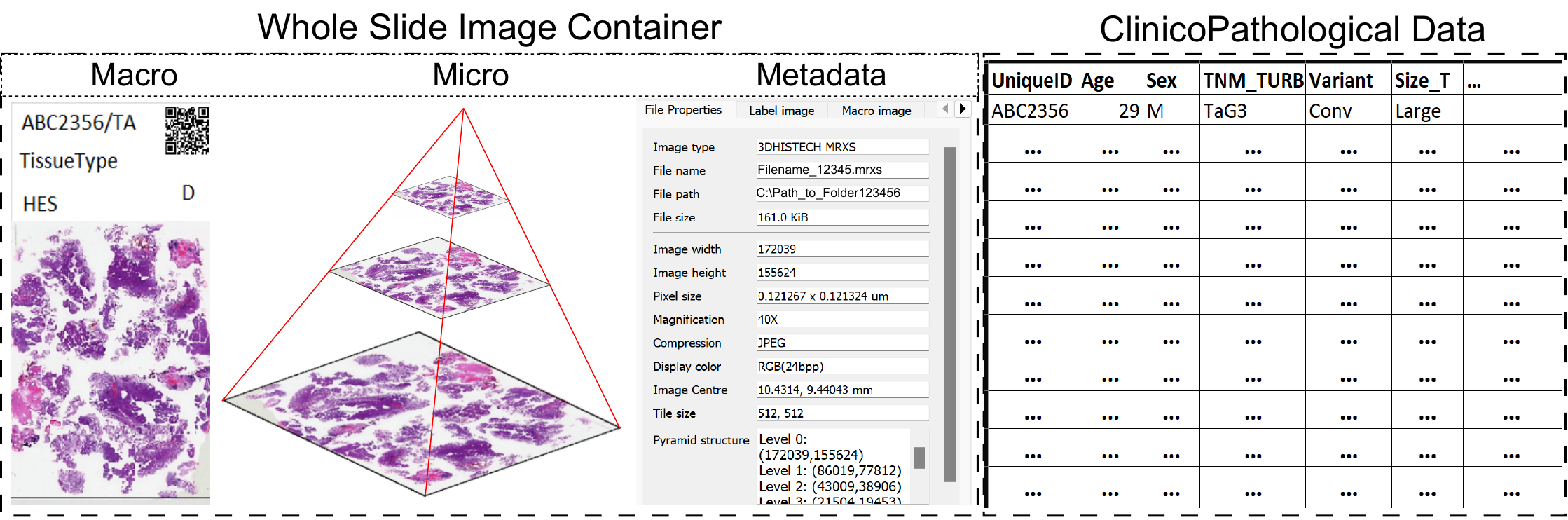}
    \caption{\textbf{A depiction of elements in histopathological data.} The whole slide image container contains macro (a glass slide with a label), micro (high-level tissue information), and metadata (technical and administrative information from the scanner). Healthcare institutions differently maintain clinicopathological data based on their primary purpose.\vspace{-0.5em}}
    \label{fig1}
\end{figure*}

%% Multiparty computing
With the advent of \emph{Multiparty Computing} (MPC), large-scale data processing is now possible by calculating common functions collaboratively, where chunks of data with multiple parties are meaningless without other pieces. In brief, secure MPC is cryptographic computing that can help bring computation to private data and can be less computationally complex than HE. Welten \etal~\cite{welten2022privacy} proposed a framework leveraging MPC for healthcare data. They aimed to establish a distributed analytics platform but did not consider AI model inversion attacks~\cite{geng2023improved}. Similarly, Federated Learning (FL) allows training AI models on premises without transferring data~\cite{tabatabaei2023wwfedcbmir}. Geng \etal~\cite{geng2021did} proposed a decentralized identity-based system for facilitating trustworthy FL using a smart contract. Their architecture concept lacked specifications for using large medical images from multiple institutions and protection of AI model weights. Even though the data is pseudonymized, there are risks of information leakage while sharing the AI model weights in FL~\cite{geng2023improved}. Nonetheless, MPC and FL can be effective collaborative research and analysis solutions. For processing very large histopathological images over distributed cloud resources, Wang \etal~\cite{wang} proposed artifact detection~\cite{kanwal2023vision} cloud-based DP service. Their methodology involved stripping metadata and other sensitive information in trusted nodes before exporting image data to external cloud resources. Their methodology utilized encryption for the chunks of sub-image locations as a preventive measure against the image-matching algorithm in case of data leakage in computing nodes.

Apart from the known drawbacks of these privacy-preserving frameworks, such as high communication costs, complexity, and loss of information, they have limitations when applied to histopathology. Identifying technical and quantitative criteria to choose a particular approach for histopathological data is challenging.

% However, direct exposure of raw data to third-party services invokes risks to patient privacy; Therefore, only trusted participants can be involved even though the medical data is pseudonymized

% An extension of cryptography-based multiparty computation methods is aimed to ensure data only leaves the individual encryption level.

% A blockchain-based plan for maintaining medical data was put up by Lee \etal~\cite{lee2021blockchain}. Only those who are authorized can access the data under the proposed system. 

\section{Histopathology: Use Case}
% Case studies demonstrating the effectiveness of the anonymization framework for whole slide images in preserving privacy and enabling medical data sharing
% Discussion of the limitations and challenges of the framework
% Discussion of the specific use case for the anonymization framework in the context of the author's research or education
% Discussion of the potential impact of the framework on the author's research or education
% \vspace{-1em}

% \begin{figure*}[ht!]
%     \centering
%     \includegraphics[width=18cm]{Paper_Figure1.pdf}
%     \caption{\textbf{A depiction of elements in histopathological data.} The whole slide image container contains macro (a glass slide with a label), micro (high-level tissue information), and metadata (technical and administrative information from the scanner). Healthcare institutions differently maintain clinicopathological data based on their primary purpose.\vspace{-0.5em}}
%     \label{fig1}
% \end{figure*}

\vspace{0.5em}
With the growing pressure from funding agencies to make medical data public for multidisciplinary research, the burden usually falls on healthcare institutions to comply with regulations. Nevertheless, privacy laws and regulations do not provide straightforward operational methods for releasing different types of medical data~\cite{el2015anonymising}. Since traditional histopathology involves preparing a glass slide and observing it under a microscope, Whole Slide Scanners (WSS) are used for digitizing, and different WSS vendors use their own proprietary format for storing histopathological images~\cite{kanwal2022devil}. Despite numerous benefits, the lack of industry-wide standardization in DP and the absence of anonymization functionality in WSS have become major obstacles to sharing histopathological data. Preserving privacy in histopathology poses unique challenges as sensitive information lies in three elements: i) clinical information (often referred to as clinicopathological data),  ii) Metadata and a macro label in WSI container, and iii) tissue image (micro), as shown in Figure~\ref{fig1}; Therefore, anonymization of histopathological data can be intricate, as a practical solution would require obfuscating identifiers in all three elements.

% \vspace{-0.5em}
\subsection{Preparing Data for Release}
Histopathological datasets can be prepared and released for public, quasi-public, and non-public use~\cite{el2015anonymising}. Public datasets are usually available to anyone with the least restriction and a high degree of anonymization. TCGA~\footnote{\url{https://www.cancer.gov/ccg/access-data}} and BreakHis~\footnote{\url{https://web.inf.ufpr.br/vri/}} are two examples of publicly available histopathological datasets. Quasi-public datasets are prepared with a relatively low degree of anonymization and prohibit researchers from contacting patients or attempting re-identification. TCGA`s controlled access tier, which includes RNA sequences, is an example of such a release. It is open to only qualified researchers under institutional data certification~\cite {tcga_policies}. Non-public datasets are usually prepared in a pseudonymized fashion with maximum data utility for collaborative uses. In all three cases, data custodians in the EU follow two procedures: i) obtaining the patient`s consent and ii) applying an appropriate anonymization or pseudonymization method.

The use of medical data for primary purposes, such as diagnosis and treatment, does not require consent from the patient. Complications like legitimate privacy concerns arise when data is used for secondary purposes such as research and education. Patient consent is usually evaluated using active and passive approaches. In active consent, a letter or request can be sent to the patient, and data sharing is put on hold before the reply arrives. In passive consent, a letter can be sent to the patient with instructions for the patient to notify only if they do not want to consent to use his/her data for research purposes. Anonymization is used for public or quasi-public datasets. While anonymizing, minimizing the probability of re-identification and retaining enough information is vital. Simpler or advanced anonymization (as described in section~\ref{relatedwork}) can be applied with different degrees for public and quasi-public releases.

Finally, the Regional Ethics Committee (REC) and Data Protection Officer (DPO) oversee consent and compliance with privacy regulations for approval prior to transferring data outside the institution. An exception might be made when the patient is deceased or while investigating a large population where the REC can exclusively approve the use of data because society has an overall huge interest in the results of a study, which outweighs the disadvantages for the patient.

% \vspace{-1em}
 \subsection{Secure Data Storage Formats}
%, making it hard to use a single scheme for all data formats of WSI. 
Establishing a single anonymization tool for all formats in DP is unfeasible due to the different structures of metadata in WSI formats, which encourages adopting a standard format. Among possible future adoptions, DICOM~\footnote{\url{https://www.dicomstandard.org/using/security}} and  OMERO~\footnote{\url{https://www.openmicroscopy.org/omero/}} formats are potential candidates as both are already being used in several medical domains. 

Digital Imaging and Communications in Medicine (DICOM) is a non-proprietary data interchange protocol that standardizes medical images and metadata for interoperability between healthcare systems. DICOM object includes specifications for describing image graphics objects and is usable with Picture Archiving and Communication Systems (PACS). Microscopy Environment (OMERO) is an open-source data management tool for exchanging microscopy images and associated metadata. OMERO is customizable, with a flexible data model for integration with other software tools. PACS and OMERO are both used in the medical field, but they are incompatible.
WSI from other data formats can be converted to DICOM or OMERO for uniform metadata fields and step towards simplifying the deletion of sensitive information in histopathological data. Interestingly, there are several open-source anonymization tools available for both DICOM and OMERO formats, such as DICOM Anonymizer~\footnote{\url{https://dicomapps.com/dicom-anonymizer/index.html}}, DICOM Cleaner~\footnote{\url{http://www.dclunie.com/pixelmed/software/webstart/DicomCleanerUsage.html}}, ARX~\footnote{\url{https://arx.deidentifier.org/anonymization-tool/risk-analysis/}}, including OMERO`s built-in anonymization features. 

% \vspace{-1em}
\subsection{Sharing Histopathological Data}
There is always a trade-off between the selection of pseudonymization and anonymization techniques based on who/where the histopathological data is being used and the level of openness required for the targeted research. The ultimate objective of sharing histopathological data for AI research is to benefit the development of Computational Pathology (CPATH) services. The following guidelines and suggestions can be considered for sharing histopathological data:

\begin{itemize}

    \item To mitigate the risks of insider threats, while medical data collection is in process, hospital databases should be encrypted, and access control mechanisms can be implemented to avoid unauthorized access.
    
     \item The REC and DPO play an important role in approving the use of medical data for multidisciplinary projects. A legally enforceable agreement between the data custodian and the data recipient must be in place to address data ownership, permitted uses, data retention, and safeguards to protect patient privacy. 
     
    \item When researchers perform analysis of the histopathological data stored on the institution's premises, organizational measures for pseudonymization are sufficient. A lightweight agreement between the data custodian and the data recipient should be signed to preclude re-identifying data subjects and/or inferring about a specific person. If AI models are trained on images and model weights are transported outside for distributed learning, the agreement may also enforce guarantees against deep leakage attacks~\cite {geng2023improved}. 
   
    \item For creating a large cohort of histopathological images from different institutions across the globe, WSIs should be transformed into a single format to apply the same degree of anonymization and harmonize the structure of clinicopathological data for more accurate analysis.
 
    \item Strong disassociation should be established while sending WSIs prepared from the same tissue sample to public and non-public datasets. Since the two datasets (with varying degrees of data utility) may be aimed at different analyses; there is a likelihood of inferring information by image-matching algorithms due to rare cancer diagnoses and mutations.
    
    \item Non-public datasets in collaborative research may not be fully anonymous; therefore, permission to use data should be time-constrained to re-assess the risk of identification with evolving data-linkage attacks. In addition, if public servers are planned for computational use, distributed histopathological image processing~\cite{wang} should be applied to avoid leakage of the entire WSI.
    
    \item Metadata, macro image label, and all quasi-identifiers should be removed from WSIs when releasing data for public use. Moreover, a risk assessment for the probability of re-identification should be performed, as it becomes impossible to call back medical data after its release.

\end{itemize}

\section{Discussion and Future Directions}
% Summary of the key findings of the research paper
% Conclusion 
% Growing research of AI in CPATH,
% Robust AI requires large datasets and troubles in applying privacy.
% understanding different techniques and definitions of privacy-related terms in researchers from different backgrounds.
% lack of consensus in adopting a single method as a standard. 
% \vspace{-0.5em}
\vspace{0.5em}
Research in DP has seen an increasing use of AI algorithms. Developing AI-based CPATH services can benefit clinical practices. However, developing robust AI algorithms requires extensive medical data collection with broad and diverse disease patterns. Though there are several pseudonymization and anonymization techniques to allow the secondary use of medical data, they introduce complications in facilitating data sharing for bio-informatics research. First, a lack of understanding of the nuances of legal and technical terminologies for researchers and educators in interdisciplinary projects induces unintentional risks of opening medical data to disclosure attacks. Secondly, there is no firm agreement on the adaptability of a single method as a standard for privacy preservation. Several articles~\cite{el2011systematic,rajendran2017study,vokinger2020lost} have also argued for the risk of re-identification of de-identified data as it becomes easy to reverse using the growing advancement of data linkage techniques~\cite{sweeney2013matching}. Also, the concept of "reasonable effort" in re-identification has been changed due to AI-powered de-anonymization techniques. These challenges have halted the adoption of a widely accepted framework for smooth data exchange.

% In histopathology case
% inferring sensitive data using WSI. 
% Implications of the research for medical data sharing and anonymization.
% Suggestions for improving the anonymization framework for whole slide images
In the histopathology domain, where the goal is to develop CPATH systems for diagnosis, prognosis, and treatment outcome prediction, the availability of WSI, metadata, and clinicopathological information is vital. Methods proposing complete anonymization of histopathological data will limit the capacity for meaningful analysis. Meanwhile, using pseudonymization as a substitute for anonymization may help develop long-term applications such as prognosis. To reduce the risks of inferring sensitive data, a new strategy is required for applied data-sharing techniques, which must incorporate organizational, legal, ethical, and technical considerations. Unfortunately, different geographies of health institutions and practices in the DP community have slowed the process of the census over a privacy-protected data exchange. One of the obvious reasons is that standards like DICOM have yet to be practically adopted in WSS, and practices for maintaining clinicopathological data vary globally across health institutions. Therefore, a new anonymization technique that creates a balance by creating \emph{pseudo-anonymized} data is needed to boost biomedical research and education.

% Future direction
% Recommendations for future research in this area
The future of medical data sharing relies heavily on the effective combination of emerging technologies in privacy-preserving frameworks. Several revolutionary concepts, such as Federated Learning (FL), differential privacy, and blockchain technology, offer promising solutions to existing challenges for AI research in DP. Although FL suggests moving computation to the data to reduce the burden of applying traditional anonymization methods, it is mainly unfeasible for healthcare institutions to own and maintain high-performance resources on their premises~\cite{wang}, and processing WSIs in a timely manner is nearly impracticable for ordinary computers. Differential privacy works by adding controlled noise to the data, ensuring that any analysis performed on the data will not reveal sensitive information. However, the computational complexity involved in rigorous mathematical modeling of the data makes it difficult to determine the appropriate amount of noise to add to the data to avoid negatively impacting the accuracy of the analysis. Blockchain technology has the potential to create a secure decentralized wallet that empowers data custodians to allow the use of data on a need-to-know basis. Transaction blocks in the blockchain offer immutable storage of access records and provide a transparent audit of data exchange. Nonetheless, If healthcare institutions adopt different blockchain platforms, then their interoperability would be a foreseen challenge.

For designing a trust-worthy data-sharing platform, the stakeholders and governance frameworks must harmonize GDPR and HIPAA regulations and adhere to ethical principles that ensure the responsible use of health data and give maximum rights to the data owner. A reliable data exchange platform with reasonable computational complexity would promote transparency and trust in hassle-free medical data sharing globally for the greater good. Anonymization is not a one-time process but an ongoing effort. As new computing technologies advance, staying updated on the latest trends and data re-identification attacks is essential while harnessing AI's potential for CPATH. The future of medical data sharing is bright as long as technological and legal developments keep \emph{striking the right balance between privacy and progress}.

% privacy-preserving research practices will foster patient trust, promote scientific collaboration, and advance biomedical knowledge.

\section*{Acknowledgment}
This research has received financial support from CLARIFY project under Marie Skłodowska-Curie Actions, grant agreement No. 860627.  The authors have no relevant financial or non-financial interests to disclose.

\bibliographystyle{IEEEtran}
\bibliography{main}

\end{document}